\documentclass{article} 
\usepackage{iclr2019_conference,times}

\usepackage{amsmath,amsfonts,bm}
\usepackage{letltxmacro}
\LetLtxMacro{\originaleqref}{\eqref}

\def\1{\bm{1}}

\def\vv{{\bm{v}}}

\DeclareMathAlphabet{\mathsfit}{\encodingdefault}{\sfdefault}{m}{sl}
\SetMathAlphabet{\mathsfit}{bold}{\encodingdefault}{\sfdefault}{bx}{n}

\newcommand{\E}{\mathbb{E}}

\newcommand{\R}{\mathbb{R}}

\newcommand{\Cov}{\mathrm{Cov}}

\DeclareMathOperator{\Tr}{Tr}

\usepackage{booktabs}
\usepackage{multirow}
\usepackage{array}
\usepackage{graphics}
\usepackage{algorithm, algpseudocode}
\usepackage{mathtools}
\usepackage{enumerate}
\usepackage{amsmath}
\usepackage{subcaption}
\usepackage{xspace}
\usepackage{wrapfig}
\usepackage{pifont}
\usepackage{xcolor}
\usepackage{multirow,graphicx}
\usepackage{float}
\usepackage{hyperref}
\usepackage{nicefrac}
\usepackage{bm}
\definecolor{mydarkblue}{rgb}{0,0.08,0.45}
\definecolor{myfavblue}{rgb}{0.1176, 0.392, 1.0}
\hypersetup{
    colorlinks=true,
    linkcolor=mydarkblue,
    citecolor=mydarkblue,
    filecolor=mydarkblue,
    urlcolor=mydarkblue}
    
\usepackage{url}
\usepackage[title]{appendix}

\newcommand{\fw}{0.31\linewidth}

\definecolor{myred}{RGB}{215,48,39}
\definecolor{mygreen}{RGB}{26,152,80}
\newcommand{\cmark}{\textcolor{mygreen}{\ding{51}}}
\newcommand{\xmark}{\textcolor{myred}{\ding{55}}}

\newcommand{\PreserveBackslash}[1]{\let\temp=\\#1\let\\=\temp}
\newcolumntype{C}[1]{>{\PreserveBackslash\centering}p{#1}}
\newcolumntype{R}[1]{>{\PreserveBackslash\raggedleft}p{#1}}
\newcolumntype{L}[1]{>{\PreserveBackslash\raggedright}p{#1}}

\newcommand{\z}{\mathbf{z}}
\newcommand{\x}{\mathbf{x}}
\newcommand{\veps}{\bm{\epsilon}}

\newcommand{\methodname}{FFJORD\xspace}
\newcommand{\JTr}{\Tr\left(\frac{\partial f}{\partial \z(t)}\right)}
\newcommand{\jtr}{\Tr\left(\nicefrac{\partial f}{\partial \z(t)}\right)}

\newcommand\0{\kern-1.2pt\vec{\kern1.2pt 0}}
\newcommand{\bigO}{\mathcal{O}}

\newcommand*\samethanks[1][\value{footnote}]{\footnotemark[#1]}

\title{\methodname: Free-form Continuous Dynamics for Scalable Reversible Generative Models}
\iclrfinalcopy
\author{\hspace{-1mm}Will Grathwohl\thanks{Equal contribution. Order determined by coin toss. \{wgrathwohl, rtqichen\}@cs.toronto.edu}\;\thanks{University of Toronto and Vector Institute}\;\thanks{OpenAI}\;, Ricky T. Q. Chen\samethanks[1]\;\samethanks[2]\;, Jesse Bettencourt\samethanks[2]\;, Ilya Sutskever\samethanks[3]\;, David Duvenaud\samethanks[2]
}

\iclrfinalcopy 
\begin{document}

\maketitle

\begin{abstract}
A promising class of generative models maps points from a simple distribution to a complex distribution through an invertible neural network.
Likelihood-based training of these models requires restricting their architectures to allow cheap computation of Jacobian determinants.
Alternatively, the Jacobian trace can be used if the transformation is specified by an ordinary differential equation.
In this paper, we use Hutchinson's trace estimator to give a scalable unbiased estimate of the log-density.
The result is a continuous-time invertible generative model with unbiased density estimation and one-pass sampling, while allowing unrestricted neural network architectures.
We demonstrate our approach on high-dimensional density estimation, image generation, and variational inference, achieving the state-of-the-art among exact likelihood methods with efficient sampling.
\end{abstract}

\section{Introduction}
\begin{wrapfigure}[23]{r}{0.4\linewidth}
    \centering
    \vspace{-0.5em}%
    \includegraphics[width=0.4\textwidth]{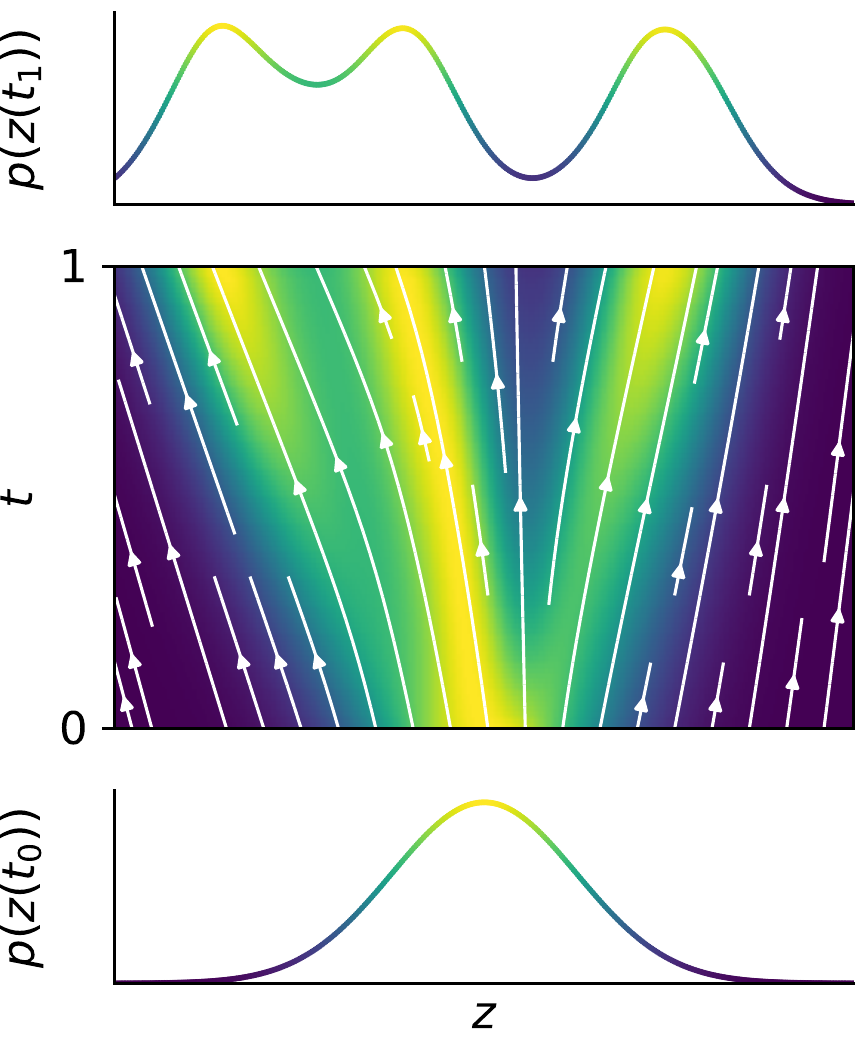}
    \vspace{-1.5em}
    \caption{\methodname{} transforms a simple base distribution at $t_0$ into the target distribution at $t_1$ by integrating over learned continuous dynamics.}
    \label{fig:toy1d}
\end{wrapfigure}
Reversible generative models use cheaply invertible neural networks to transform samples from a fixed base distribution.
Examples include NICE~\citep{dinh2014nice}, Real NVP~\citep{dinh2016density}, and Glow~\citep{kingma2018glow}. 
These models are easy to sample from, and can be trained by maximum likelihood using the change of variables formula.
However, this requires placing awkward restrictions on their architectures, such as partitioning dimensions or using rank one weight matrices, in order to avoid an $\mathcal{O}(D^3)$ cost determinant computation.

Recently, \citet{chen2018neural} introduced a continuous-time analog of normalizing flows, defining the mapping from latent variables to data using ordinary differential equations (ODEs). 
In their model, the likelihood can be computed using relatively cheap trace operations.
A more flexible, but still restricted, family of network architectures can be used to avoid this $\mathcal{O}(D^2)$ time cost. 

Extending this work, we introduce an unbiased stochastic estimator of the likelihood that has $\mathcal{O}(D)$ time cost, allowing completely unrestricted architectures.
Furthermore, we have implemented GPU-based adaptive ODE solvers to train and evaluate these models on modern hardware.
We call our approach Free-form Jacobian of Reversible Dynamics (\methodname{}).

\section{Background: Generative Models and Change of Variables}\label{sec:cov_models}
In contrast to directly parameterizing a normalized distribution~\citep{oord2016pixel,germain2015made}, the change of variables formula allows one to specify a complex normalized distribution implicitly by warping a normalized base distribution $p_\z(\z)$ through an invertible function ${f: \R^D \rightarrow \R^D}$.
Given a random variable $\z \sim p_\z(\z)$ the log density of $\x = f(\z)$ follows
\begin{equation}\label{eq:discrete_cov}
    \log p_\x(\x) = \log p_\z(\z) - \log \det \left| \frac{\partial f(\z)}{\partial \z} \right|
\end{equation}
where $\nicefrac{\partial f(\z)}{\partial \z}$ is the Jacobian of $f$.
In general, computation of the log determinant has a time cost of $\bigO(D^3)$.
Much work have gone into using restricted neural network architectures to make computing the Jacobian determinant more tractable. These approaches broadly fall into three categories:
\begin{enumerate}[\indent {}]
    \item \textbf{Normalizing flows.} By restricting the functional form of $f$, various determinant identities can be exploited \citep{rezende2015variational,berg2018sylvester}.
    These models cannot be trained directly on data and be able to sample because they do not have a tractable analytic inverse $f^{-1}$ but have been shown to be useful in representing the approximate posterior for variational inference~\citep{kingma2013auto}.
    \item \textbf{Autoregressive transformations.} By using an autoregressive model and specifying an ordering in the dimensions, the Jacobian of $f$ is enforced to be lower triangular~\citep{kingma2016improved,oliva2018transformation}.
    These models excel at density estimation for tabular datasets~\citep{papamakarios2017masked}, but require $D$ sequential evaluations of $f$ to invert, which is prohibitive when $D$ is large.
    \item \textbf{Partitioned transformations.} Partitioning the dimensions and using affine transformations makes the determinant of the Jacobian cheap to compute, and the inverse $f^{-1}$  computable with the same cost as $f$~\citep{dinh2014nice,dinh2016density}.
    This method allows the use of convolutional architectures, excelling at density estimation for image data~\citep{dinh2016density,kingma2018glow}.
\end{enumerate}
Throughout this work, we refer to \emph{reversible generative models} those that use the change of variables to transform a base distribution to the model distribution while maintaining both efficient density estimation and efficient sampling capabilities using a single pass of the model.

\subsection{Other Generative Models}
There exist several approaches to generative modeling approaches which don't use the change of variables equation for training.
Generative adversarial networks (GANs)~\citep{goodfellow2014generative} use large, unrestricted neural networks to transform samples from a fixed base distribution.
Lacking a closed-form likelihood, an auxiliary discriminator model must be trained to estimate various divergences in order to provide a training signal.
Autoregressive models~\citep{germain2015made,oord2016pixel} directly specify the joint distribution $p(\x)$ as a sequence of explicit conditional distributions using the product rule.
These models require at least $\bigO(D)$ evaluations to sample from.
Variational autoencoders (VAEs)~\cite{kingma2013auto} use an unrestricted architecture to explicitly specify the conditional likelihood $p(x|z)$, but can only efficiently provide a lower bound on the marginal likelihood $p(x)$.

\begin{table}\centering
\renewcommand{\arraystretch}{1.3}
\begin{tabular}{@{} c | l | C{1.5cm} C{1.5cm} C{1.7cm} C{1.5cm} @{}} 
\multicolumn{2}{c|}{Method} & Train on data & One-pass Sampling & Exact log-likelihood & Free-form Jacobian \\
\midrule
\multirow{3}{*}{\rotatebox[origin=c]{90}{\parbox[c]{1.6cm}{\centering \footnotesize }}} & Variational Autoencoders & \cmark & \cmark & \xmark & \cmark \\
 & Generative Adversarial Nets & \cmark & \cmark & \xmark & \cmark \\
 & Likelihood-based Autoregressive & \cmark & \xmark & \cmark & \xmark \\
\midrule
\multirow{4}{*}{\rotatebox[origin=c]{90}{\parbox[c]{2cm}{\centering \footnotesize Change of Variables}}} & Normalizing Flows & \xmark & \cmark & \cmark & \xmark \\
 & Reverse-NF, MAF, TAN & \cmark & \xmark & \cmark & \xmark \\
 & NICE, Real NVP, Glow, Planar CNF & \cmark & \cmark & \cmark & \xmark \\
 & \textbf{\methodname} & \cmark & \cmark & \cmark & \cmark
\end{tabular}
\caption{A comparison of the abilities of generative modeling approaches.}
\label{tab:comparison}
\end{table}

\subsection{Continuous Normalizing Flows}
\citet{chen2018neural} define a generative model for data $\x \in \R^D$ similar to those based on \eqref{eq:discrete_cov} which replaces the warping function with an integral of continuous-time dynamics.
The generative process works by first sampling from a base distribution $\z_0 \sim p_{z_0}(\z_0)$.
Then, given an ODE defined by the parametric function $f(\z(t), t; \theta)$, we solve the initial value problem $\z(t_0) = \z_0, \nicefrac{\partial \z(t)}{\partial t} = f(\z(t), t; \theta)$ to obtain $\z(t_1)$ which constitutes our observable data. 
These models are called Continous Normalizing Flows (CNF).
The change in log-density under this model follows a second differential equation, called the \emph{instantaneous change of variables} formula:~\citep{chen2018neural},
\begin{equation}\label{eq:cont-cov}
\frac{\partial \log p(\z(t)) }{\partial t} = -\JTr.
\end{equation}
%
We can compute total change in log-density by integrating across time:
\begin{equation}\label{eq:cont_cov_full}
    \log p(\z(t_1)) = \log p(\z(t_0)) - \int_{t_0}^{t_1} \JTr \;dt.
\end{equation}
Given a datapoint $\x$, we can compute both the point $\z_0$ which generates $\x$, as well as $\log p(\x)$ under the model by solving the initial value problem:
\begin{align}\label{eq:reverseivp}
    \underbrace{\begin{bmatrix}
    \z_0 \\ \log p(\x) - \log p_{z_0}(\z_0)
    \end{bmatrix}}_{\textnormal{solutions}} = \underbrace{\int_{t_1}^{t_0}
    \begin{bmatrix}
    f(\z(t), t; \theta) \\ - \JTr
    \end{bmatrix}dt}_{\textnormal{dynamics}},\qquad
    \underbrace{\begin{bmatrix}
    \z(t_1) \\ \log p(\x) - \log p(\z(t_1))
    \end{bmatrix} = 
    \begin{bmatrix}
    \x \\ 0
    \end{bmatrix}}_{\textnormal{initial values}}
\end{align}
which integrates the combined dynamics of $z(t)$ and the log-density of the sample backwards in time from $t_1$ to $t_0$.
We can then compute $\log p(\x)$ using the solution of \eqref{eq:reverseivp} and adding $\log p_{z_0}(\z_0)$.
The existence and uniqueness of \eqref{eq:reverseivp} require that $f$ and its first derivatives be Lipschitz continuous~\citep{khalil2002nonlinear}, which can be satisfied in practice using neural networks with smooth Lipschitz activations.

\subsubsection{Backpropagating through ODE Solutions with the Adjoint Method}
CNFs are trained to maximize \eqref{eq:cont_cov_full}. This objective involves the solution to an initial value problem with dynamics parameterized by $\theta$. For any scalar loss function which operates on the solution to an initial value problem
\begin{align}\label{eq:loss}
    L(\z(t_1)) = L\left(\int_{t_0}^{t_1} f(\z(t), t; \theta)dt\right)
\end{align}
then \citet{pontryagin2018mathematical} shows that its derivative takes the form of another initial value problem
\begin{align}\label{eq:adjoint}
    \frac{dL}{d\theta} = -\int_{t_1}^{t_0}\left(\frac{\partial L}{\partial \z(t)}\right)^T  \frac{\partial f(\z(t), t; \theta)}{\partial \theta} dt.
\end{align}
The quantity $-\nicefrac{\partial L}{\partial \z(t)}$ is known as the adjoint state of the ODE. \citet{chen2018neural} use a black-box ODE solver to compute $\z(t_1)$, and then another call to a solver to compute \eqref{eq:adjoint} with the initial value $\nicefrac{\partial L}{\partial \z(t_1)}$.
This approach is a continuous-time analog to the backpropgation algorithm~\citep{rumelhart1986learning,andersson2013general} and can be combined with gradient-based optimization methods to fit the parameters $\theta$.

\section{Scalable density evaluation with unrestricted architectures}

Switching from discrete-time dynamics to continuous-time dynamics reduces the primary computational bottleneck of normalizing flows from $\mathcal{O}(D^3)$ to $\mathcal{O}(D^2)$, at the cost of introducing a numerical ODE solver.
This allows the use of more expressive architectures.
For example, each layer of the original normalizing flows model of \citet{rezende2015variational} is a one-layer neural network with only a single hidden unit.
In contrast, the instantaneous transformation used in planar continuous normalizing flows~\citep{chen2018neural} is a one-layer neural network with many hidden units.
In this section, we construct an unbiased estimate of the log-density with $\mathcal{O}(D)$ cost, allowing completely unrestricted neural network architectures to be used.

\subsection{Unbiased Linear-time Log-Density Estimation}

In general, computing $\jtr$ exactly costs $\mathcal{O}(D^2)$, or approximately the same cost as $D$ evaluations of $f$, since each entry of the diagonal of the Jacobian requires computing a separate derivative of $f$.
However, there are two tricks that can help.
First, vector-Jacobian products $\vv^T \frac{\partial f}{\partial \z}$ can be computed for approximately the same cost as evaluating $f$, using reverse-mode automatic differentiation.
Second, we can get an unbiased estimate of the trace of a matrix by taking a double product of that matrix with a noise vector:
\begin{align}\label{eq:trace_estimator}
\Tr(A) = E_{p(\veps)}[\veps^T A \veps].
\end{align}
The above equation holds for any $D$-by-$D$ matrix $A$ and distribution $p(\veps)$ over $D$-dimensional vectors such that $\E[\veps] = 0$ and $\Cov(\veps) = I$.
The Monte Carlo estimator derived from \eqref{eq:trace_estimator} is known as the Hutchinson's trace estimator~\citep{hutchinson1989trace,adams2018trace}.

To keep the dynamics deterministic within each call to the ODE solver, we can use a fixed noise vector $\veps$ for the duration of each solve without introducing bias:
%
\begin{align}\label{eq:stochastic_ode}
    \log p(\z(t_1)) & = \log p(\z(t_0)) - \int_{t_0}^{t_1} \JTr dt\nonumber\\
              &= \log p(\z(t_0)) - \int_{t_0}^{t_1} \E_{p(\veps)}\left[\veps^T \frac{\partial f}{\partial \z(t)}\veps\right]dt \nonumber\\
              &= \log p(\z(t_0)) - \E_{p(\veps)}\left[\int_{t_0}^{t_1} \veps^T \frac{\partial f}{\partial \z(t)}\veps dt \right]
\end{align}
%
Typical choices of $p(\veps)$ are a standard Gaussian or Rademacher distribution~\citep{hutchinson1989trace}. 

\subsubsection{Reducing Variance with Bottleneck Capacity}

Often, there exist bottlenecks in the architecture of the dynamics network, i.e. hidden layers whose width $H$ is smaller than the dimensions of the input $D$.
In such cases, we can reduce the variance of Hutchinson's estimator by using the cyclic property of trace.
Since the variance of the estimator for $\Tr(A)$ grows asymptotic to $||A||_F^2$~\citep{hutchinson1989trace}, we suspect that having fewer dimensions should help reduce variance.
If we view view the dynamics as a composition of two functions $f = g \circ h(\z)$ then we observe
\begin{equation}\label{eq:bottleneck_trick}
    \Tr\underbrace{\left( \frac{\partial f}{\partial \z} \right)}_{D \times D} = \Tr\underbrace{\left( \frac{\partial g}{\partial h}\frac{\partial h}{\partial \z} \right)}_{D \times D} = \Tr\underbrace{ \left( \frac{\partial h}{\partial \z}\frac{\partial g}{\partial h} \right)}_{H \times H} = \E_{p(\veps)}\left[ \veps^T \frac{\partial h}{\partial \z}\frac{\partial g}{\partial h} \veps\right].
\end{equation}
When $f$ has multiple hidden layers, we choose $H$ to be the smallest dimension.
This bottleneck trick can reduce the norm of the matrix which may also help reduce the variance of the trace estimator.

\subsection{\methodname: A Continuous-time Reversible Generative Model}
Our complete method uses the dynamics defined in \eqref{eq:cont-cov} and the efficient log-likelihood estimator of \eqref{eq:stochastic_ode} to produce the first scalable and reversible generative model with an unconstrained Jacobian, leading to the name Free-Form Jacobian of Reversible Dyanamics (\methodname).
Pseudo-code of our method is given in Algorithm \ref{alg:ffjord}, and Table~\ref{tab:comparison} summarizes the capabilities of our model compared to previous work.

\begin{algorithm}
\caption{Unbiased stochastic log-density estimation using the \methodname model}
\label{alg:ffjord}
\begin{algorithmic}
\Require dynamics $f_\theta$, start time $t_0$, stop time $t_1$, minibatch of samples $\x$.

\State $\veps \leftarrow$ sample\_unit\_variance($\x$.shape) \Comment{Sample $\veps$ outside of the integral}
\Function{$f_{aug}$}{$[\z_t, \log p_t], t$}: \Comment{Augment $f$ with log-density dynamics.}
\State $f_t \leftarrow f_\theta(\z(t), t)$ \Comment{Evaluate dynamics}
\State $g \leftarrow \veps^T\frac{\partial f}{\partial \z}\big|_{\z(t)}$ \Comment{Compute vector-Jacobian product with automatic differentiation}
\State $\widetilde{\textnormal{Tr}} = \text{matrix\_multiply}(g, \veps$) \Comment{Unbiased estimate of $\Tr(\frac{\partial f}{\partial \z}) \text{ with } \veps^T\frac{\partial f}{\partial \z}\veps$}
\State \textbf{return} $[f_t, -\widetilde{\textnormal{Tr}}]$ \Comment{Concatenate dynamics of state and log-density}
\EndFunction 
\State $[\z, \Delta_{logp}] \leftarrow$ \textnormal{odeint}($f_{aug}$, $[\x, \0]$, $t_0$, $t_1$) \Comment{Solve the ODE, ie. $\int_{t_0}^{t_1} f_{aug}([\z(t), \log p(\z(t))], t)\;dt$}
\State $\log \hat{p}(\x) \leftarrow \log p_{\z_0}(\z)$ - $\Delta_{logp}$ 
\Comment{Add change in log-density}\\
\Return $\log \hat{p}(\x)$
\end{algorithmic}
\end{algorithm}
Assuming the cost of evaluating $f$ is on the order of $\bigO(DH)$ where $D$ is the dimensionality of the data and $H$ is the size of the largest hidden dimension in $f$, then the cost of computing the likelihood in models which stack transformations that exploit \eqref{eq:discrete_cov} is $\bigO((DH + D^3)L)$ where $L$ is the number of transformations used.
For CNF, this reduces to $\bigO((DH + D^2)\hat{L})$ for CNFs, where $\hat{L}$ is the number of evaluations of $f$ used by the ODE solver.
With \methodname, this reduces further to $\bigO((DH + D)\hat{L})$.

\pagebreak
\section{Experiments}
\begin{wrapfigure}[21]{r}{0.5\linewidth}
    \vspace{-11mm}
    \centering
    \begin{subfigure}[b]{\fw}
    \centering
    \caption*{Data}
    \vspace{-2mm}
    \includegraphics[width=\linewidth]{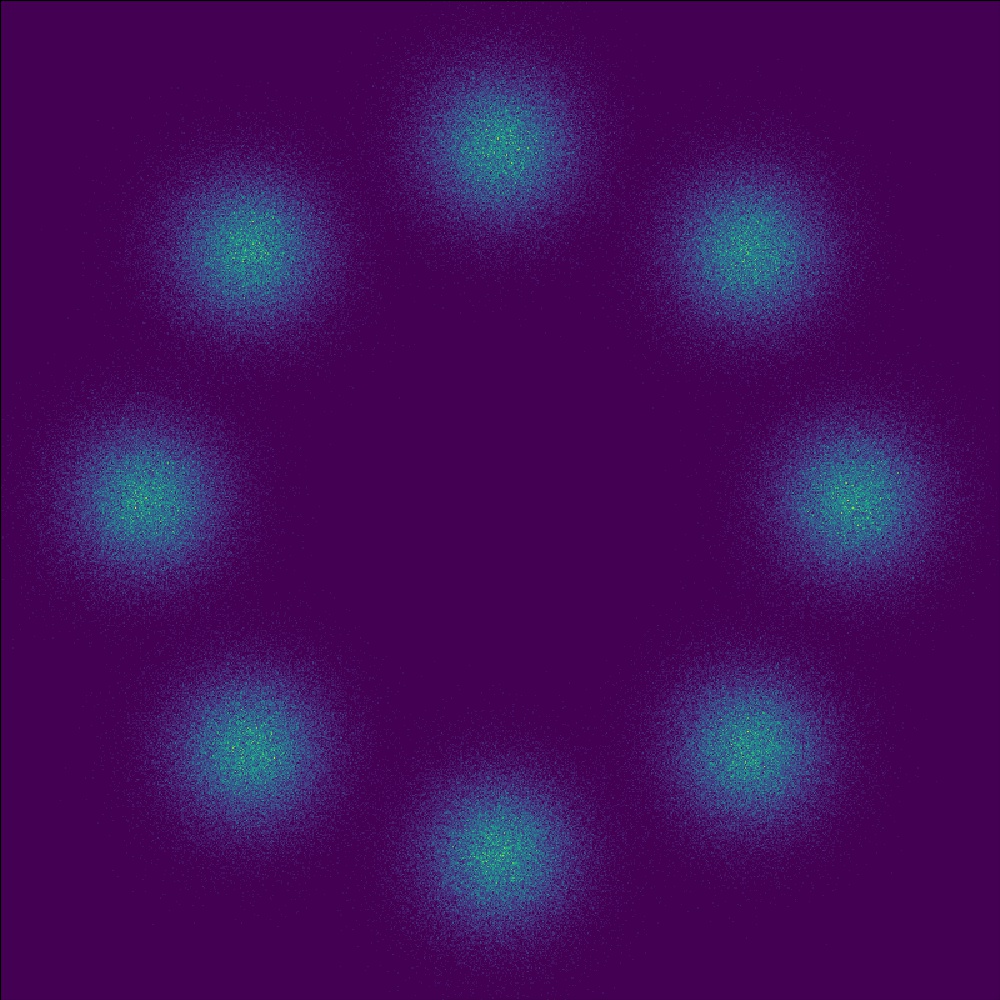}
    \end{subfigure}%
    \hspace{1mm}%
    \begin{subfigure}[b]{\fw}
    \centering
    \caption*{Glow}
    \vspace{-2mm}
    \includegraphics[width=\linewidth]{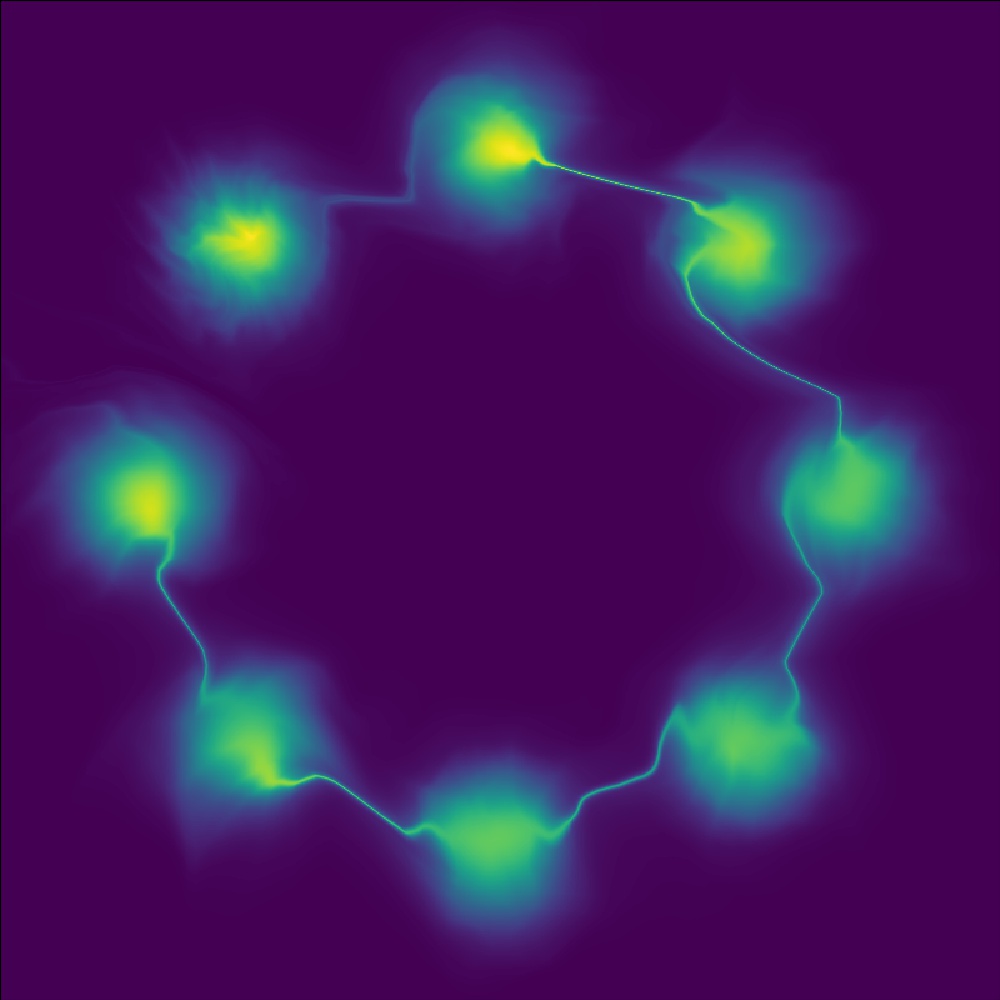}
    \end{subfigure}%
    \hspace{1mm}%
    \begin{subfigure}[b]{\fw}
    \centering
    \caption*{\methodname}
    \vspace{-2mm}
    \includegraphics[width=\linewidth]{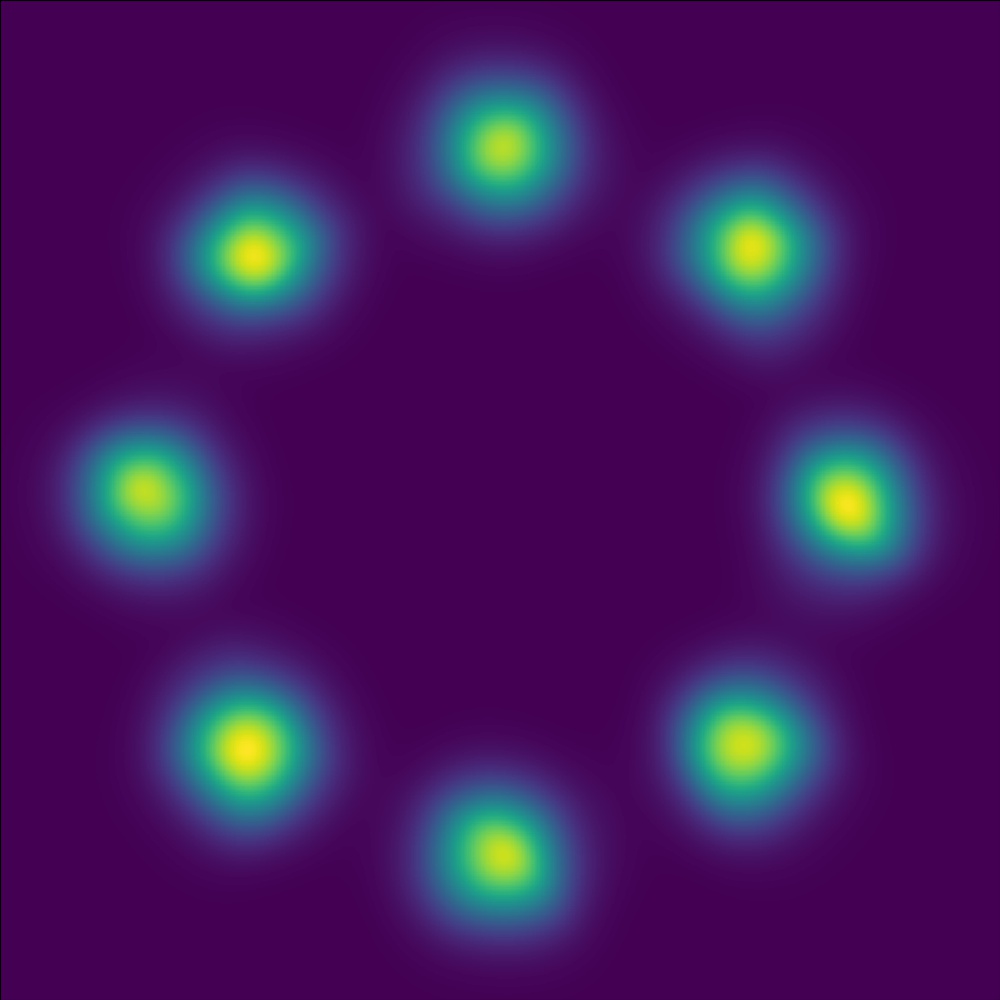}
    \end{subfigure}%

    \vspace{1em}
    \begin{subfigure}[b]{\fw}
    \centering
    \vspace{-2mm}
    \includegraphics[width=\linewidth]{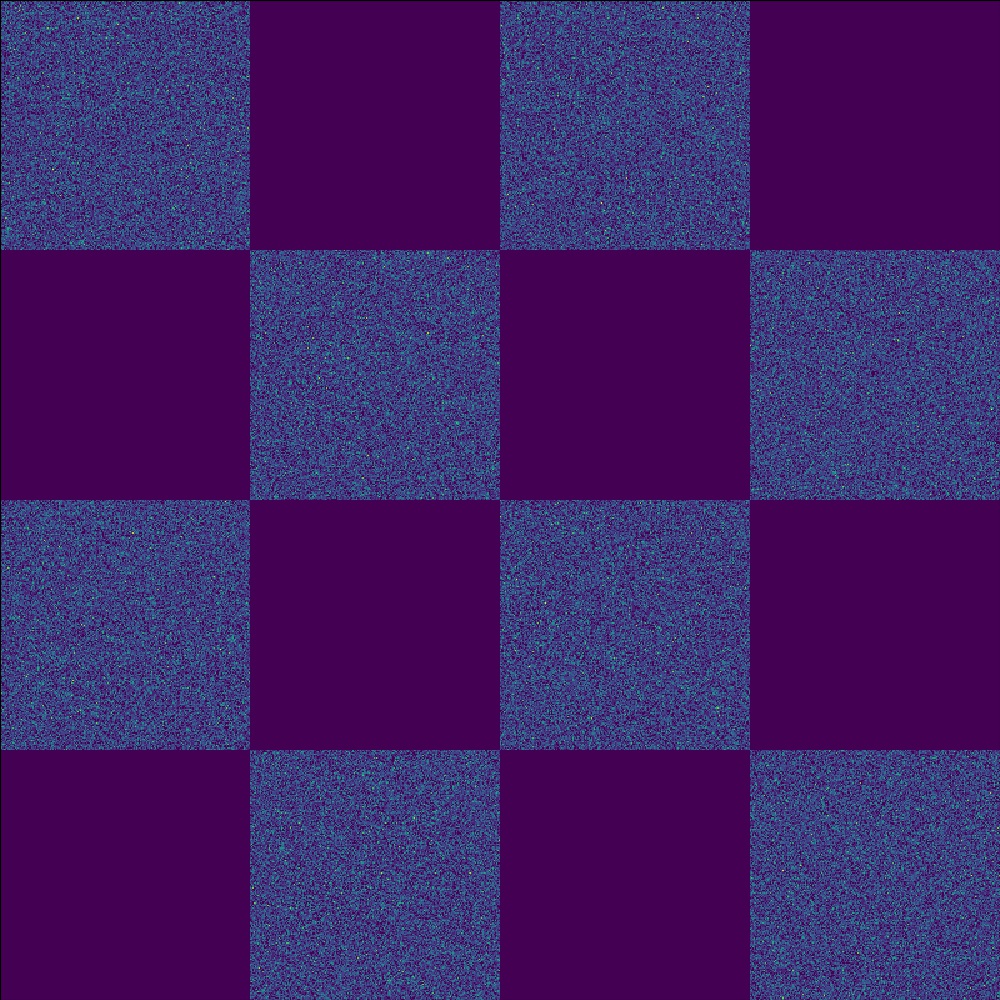}
    \end{subfigure}%
    \hspace{1mm}%
    \begin{subfigure}[b]{\fw}
    \centering
    \vspace{-2mm}
    \includegraphics[width=\linewidth]{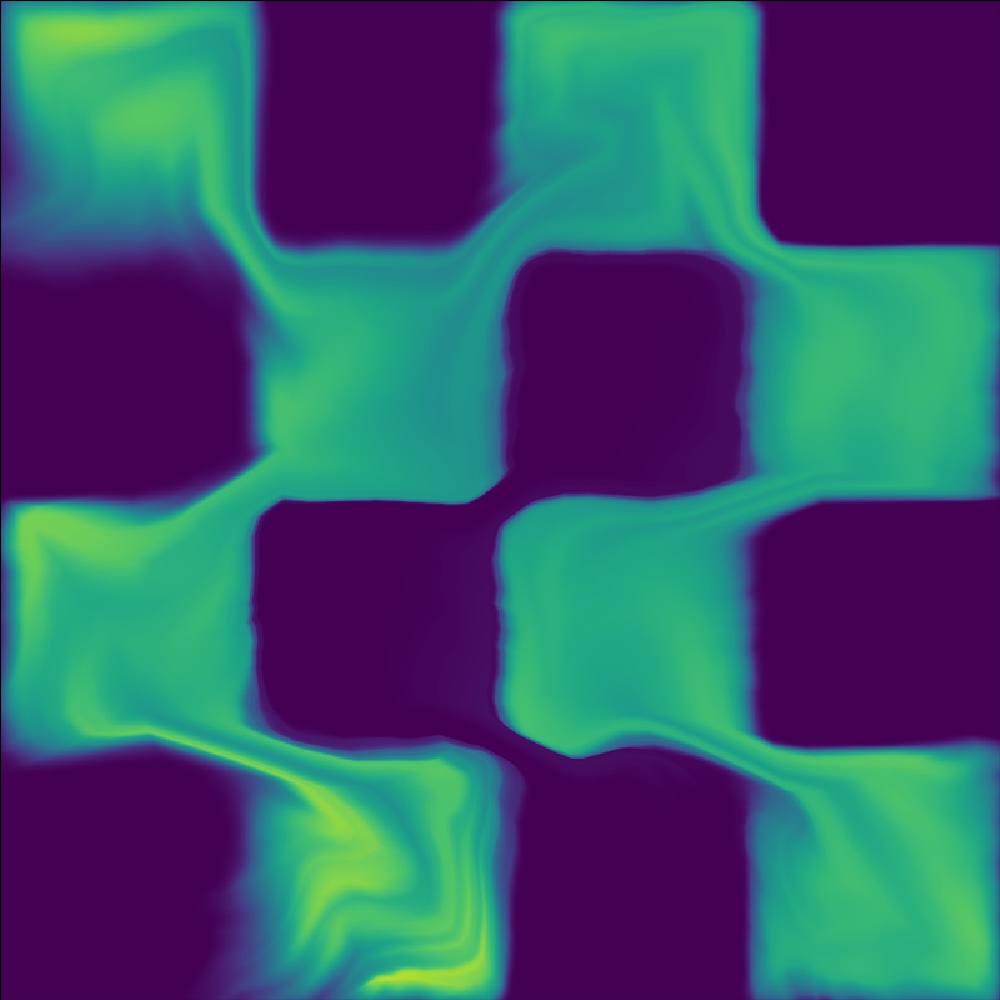}
    \end{subfigure}%
    \hspace{1mm}%
    \begin{subfigure}[b]{\fw}
    \centering
    \vspace{-2mm}
    \includegraphics[width=\linewidth]{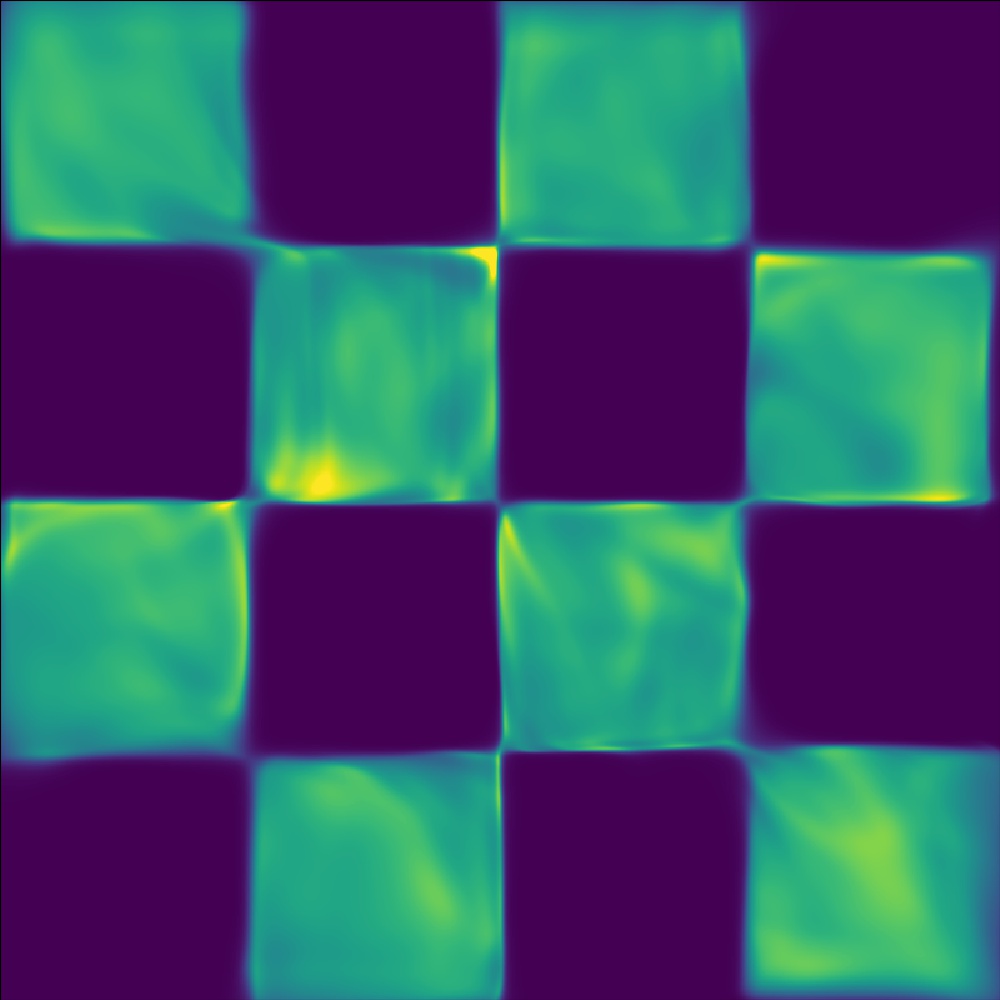}
    \end{subfigure}%
    
    \vspace{1em}
    \begin{subfigure}[b]{\fw}
    \centering
    \vspace{-2mm}
    \includegraphics[width=\linewidth]{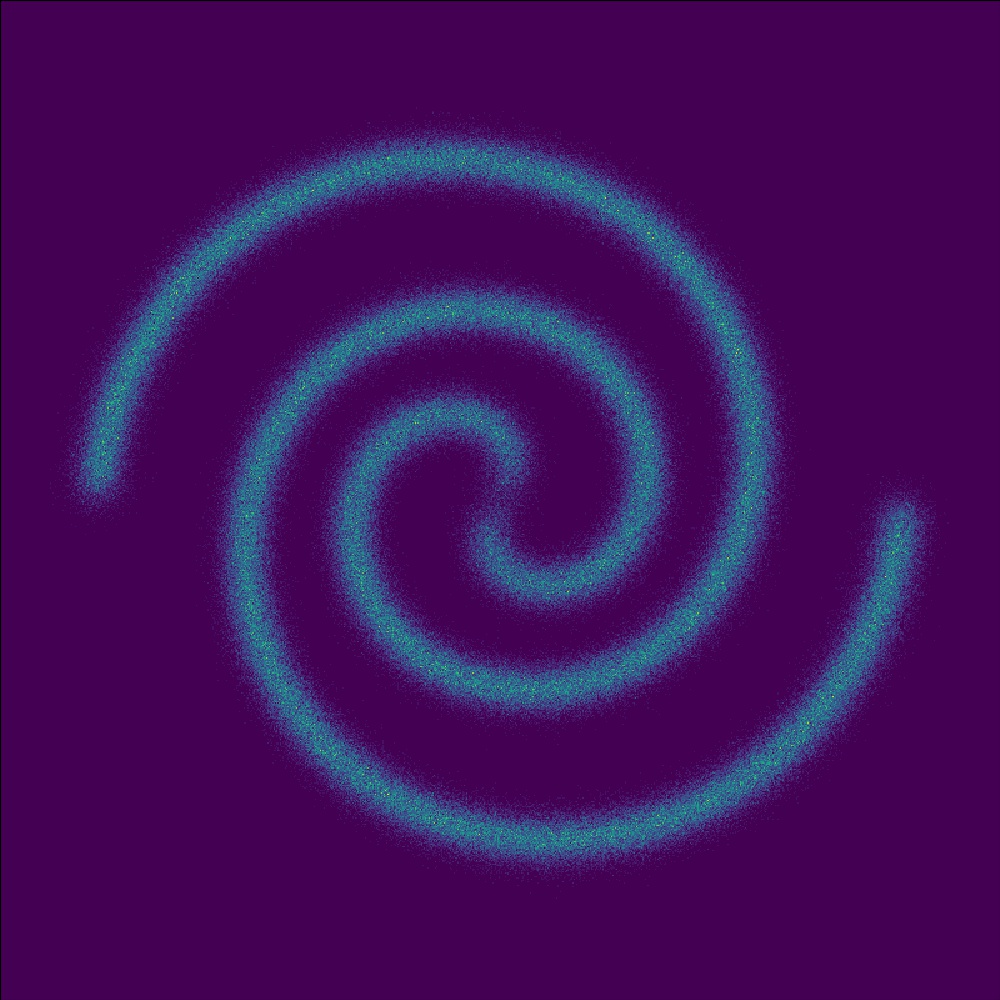}
    \end{subfigure}%
    \hspace{1mm}%
    \begin{subfigure}[b]{\fw}
    \centering
    \vspace{-2mm}
    \includegraphics[width=\linewidth]{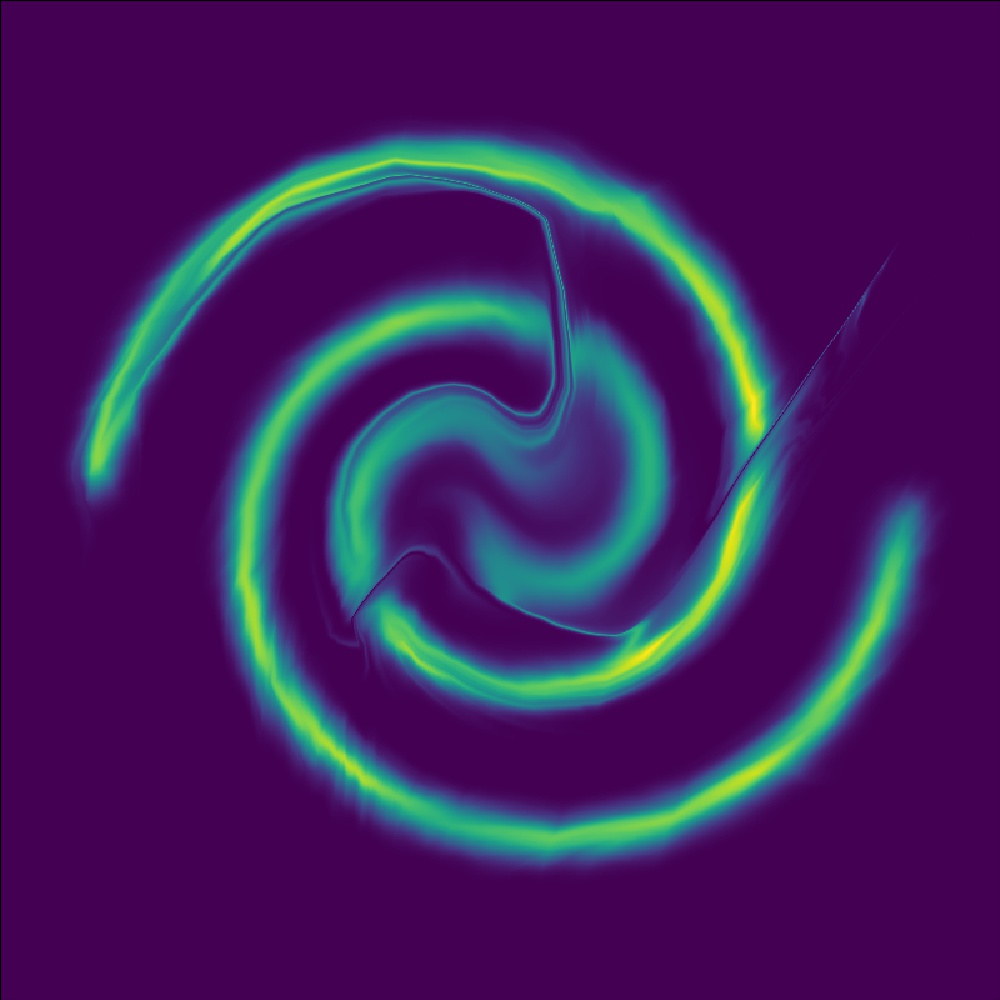}
    \end{subfigure}%
    \hspace{1mm}%
    \begin{subfigure}[b]{\fw}
    \centering
    \vspace{-2mm}
    \includegraphics[width=\linewidth]{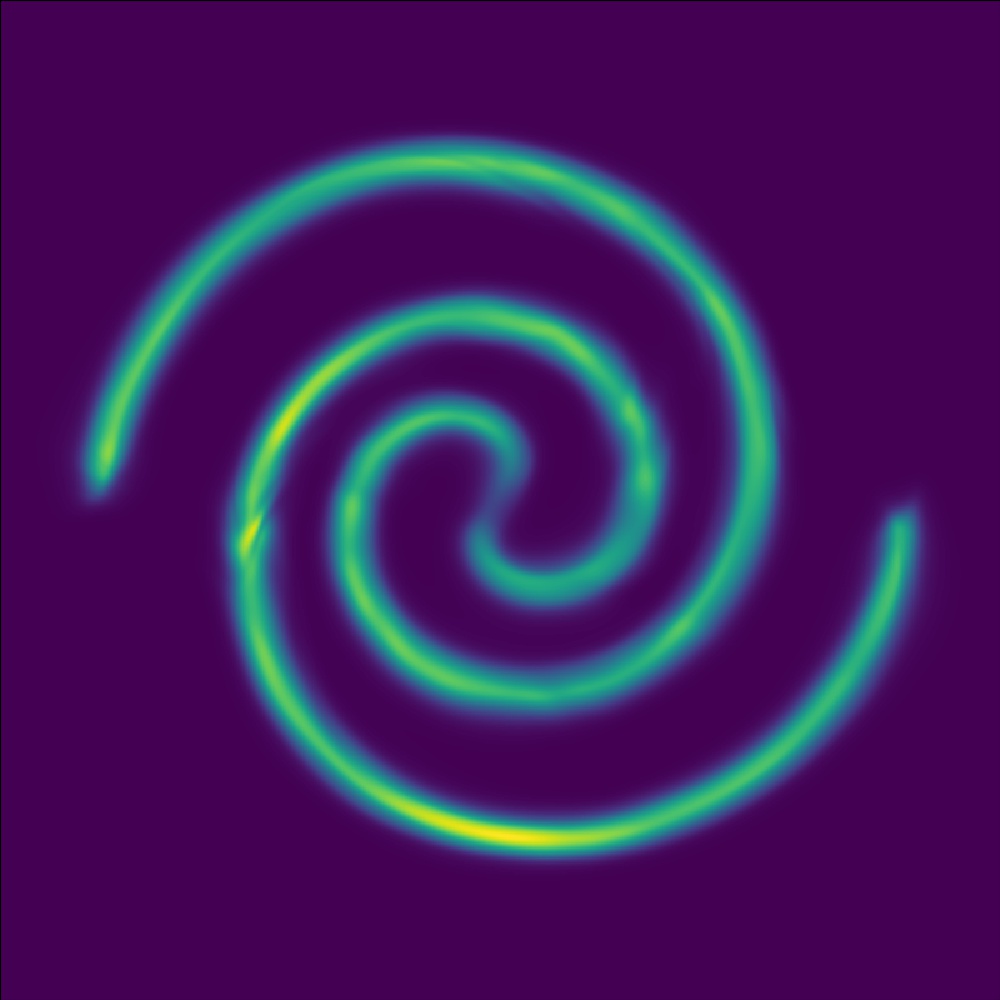}
    \end{subfigure}\\
    
    \caption{Comparison of trained \methodname and Glow models on 2-dimensional distributions including multi-modal and discontinuous densities.}
    \label{fig:2d}
\end{wrapfigure}
We demonstrate the power of \methodname on a variety of density estimation tasks as well as approximate inference within variational autoencoders \citep{kingma2013auto}.
Experiments were conducted using a suite of GPU-based ODE-solvers and an implementation of the adjoint method for backpropagation~\footnote{We plan on releasing the full code, including our GPU-based implementation of ODE solvers and the adjoint method, upon publication.}.
In all experiments the Runge-Kutta 4(5) algorithm with the tableau from \citet{shampine1986some} was used to solve the ODEs. We ensure tolerance is set low enough so numerical error is negligible; see Appendix \ref{sec:numerical_error}.

We used Hutchinson's trace estimator \eqref{eq:trace_estimator} during training and the exact trace when reporting test results.
This was done in all experiments except for our density estimation models trained on MNIST and CIFAR10 where computing the exact Jacobian trace was not computationally feasible.
There, we observed that the variance of the log-likelihood over the validation set induced by the trace estimator is less than $10^{-4}$.

The dynamics of \methodname are defined by a neural network $f$ which takes as input the current state $\z(t) \in \R^D$ and the current time $t \in \R$.
We experimented with several ways to incorporate $t$ as an input to $f$, such as hyper-networks, but found that simply concatenating $t$ on to $\z(t)$ at the input to every layer worked well and was used in all of our experiments. 

\subsection{Density Estimation On Toy 2D Data}

We first train on 2 dimensional data to visualize the model and the learned dynamics.\footnote{Videos of the learned dynamics can be found at \href{https://imgur.com/a/Rtr3Mbq}{https://imgur.com/a/Rtr3Mbq}.}
In Figure \ref{fig:2d}, we show that by warping a simple isotropic Gaussian, \methodname can fit both \emph{multi-modal} and even \emph{discontinuous} distributions.
The number of evaluations of the ODE solver is roughly 70-100 on all datasets, so we compare against a Glow model with 100 discrete layers.

The learned distributions of both \methodname and Glow can be seen in Figure \ref{fig:2d}.
Interestingly, we find that Glow learns to stretch the single mode base distribution into multiple modes but has trouble modeling the areas of low probability between disconnected regions.
In contrast, \methodname is capable of modeling disconnected modes and can also learn convincing approximations of discontinuous density functions (middle row in Figure \ref{fig:2d}).

\subsection{Density Estimation on Real Data}
We perform density estimation on five tabular datasets preprocessed as in \citet{papamakarios2017masked} and two image datasets; MNIST and CIFAR10.
On the tabular datasets, \methodname performs the best out of reversible models by a wide margin but is outperformed by recent autoregressive models.
Of those, \methodname outperforms MAF~\citep{papamakarios2017masked} on all but one dataset and manages to outperform TAN~\cite{oliva2018transformation} on the MINIBOONE dataset. 
These models require $\bigO(D)$ sequential computations to sample from while the best performing method, MAF-DDSF~\citep{huang2018neural}, cannot be sampled from analytically. 

\begin{figure}
    \centering
    \begin{subfigure}[c]{0.09\linewidth}
    Samples
    \end{subfigure}
    \begin{subfigure}[c]{.44\linewidth}
    \centering
    \includegraphics[width=\linewidth]{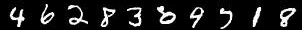}
    \end{subfigure}%
    \hfill%
    \begin{subfigure}[c]{.44\linewidth}
    \centering
    \includegraphics[width=\linewidth]{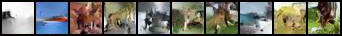}
    \end{subfigure}\\
    
    \begin{subfigure}[c]{0.09\linewidth}
    Data
    \end{subfigure}
    \begin{subfigure}[c]{.44\linewidth}
    \centering
    \includegraphics[width=\linewidth]{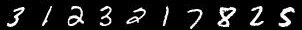}
    \end{subfigure}%
    \hfill%
    \begin{subfigure}[c]{.44\linewidth}
    \centering
    \includegraphics[width=\linewidth]{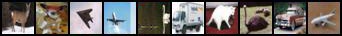}
    \end{subfigure}
    \caption{Samples and data from our image models. MNIST on left, CIFAR10 on right.}
    \label{fig:imgen_cropped}
\end{figure}

On MNIST we find that \methodname can model the data as well as Glow and Real NVP by integrating a single flow defined by one neural network.
This is in contrast to Glow and Real NVP which must compose many flows together to achieve similar performance. When we use multiple flows in a multiscale architecture (like those used by Glow and Real NVP) we obtain better performance on MNIST and comparable performance to Glow on CIFAR10.
Notably, \methodname is able to achieve this performance while \emph{using less than 2\% as many parameters} as Glow.
We also note that Glow uses a learned base distribution whereas \methodname and Real NVP use a fixed Gaussian. 
A summary of our results on density estimation can be found in Table \ref{tab:density} and samples can be seen in Figure \ref{fig:imgen_cropped}. Full details on architectures used, our experimental procedure, and additional samples can be found in Appendix \ref{sec:densityexperiment}.

In general, our approach is slower than competing methods, but we find the memory-efficiency of the adjoint method allows us to use much larger batch sizes than those methods. 
On the tabular datasets we used a batch sizes up to 10,000 and on the image datasets we used a batch size of 900. 

\begin{table}\centering
\setlength{\tabcolsep}{4pt}
\renewcommand{\arraystretch}{1.3}
\resizebox{\columnwidth}{!}{%
\begin{tabular}{@{}l|ccccc|cc@{}}
& POWER & GAS & HEPMASS & MINIBOONE & BSDS300 & MNIST & CIFAR10 \\
\midrule
Real NVP & -0.17 & -8.33 & 18.71 & 13.55 & -153.28 & 1.06* & 3.49* \\
Glow & -0.17 & -8.15 & 18.92 & 11.35 & -155.07 & 1.05* & \textbf{3.35}* \\
\methodname & \textbf{-0.46} & \textbf{-8.59} & \textbf{14.92} & \textbf{10.43} & \textbf{-157.40} & \textbf{0.99*} (1.05$^\dagger$) & 3.40* \\
\midrule
MADE & 3.08 & -3.56 & 20.98 & 15.59 & -148.85 & 2.04 & 5.67 \\
MAF  & -0.24 & -10.08 & 17.70 & 11.75 & -155.69 & 1.89 & 4.31 \\
TAN & -0.48 & -11.19 & 15.12 & 11.01 & -157.03 & - & - \\
MAF-DDSF & -0.62 & -11.96 & 15.09 & 8.86 & -157.73 & - & -
\end{tabular}
}
\caption{Negative log-likehood on test data for density estimation models; \textbf{lower is better}. In nats for tabular data and bits/dim for MNIST and CIFAR10. *Results use multi-scale convolutional architectures. $^\dagger$Results use a single flow with a convolutional encoder-decoder architecture.}
\label{tab:density}
\end{table}

\begin{table}[h]\centering
\renewcommand{\arraystretch}{1.3}
\begin{tabular}{@{}l|cccc@{}}
& MNIST & Omniglot & Frey Faces & Caltech Silhouettes \\
\midrule
No Flow     & $86.55 \pm .06$ & $104.28 \pm .39$ & $4.53 \pm .02$          & $110.80 \pm .46$ \\
Planar      & $86.06 \pm .31$ & $102.65 \pm .42$ & $4.40 \pm .06$ & $109.66 \pm .42$ \\
IAF         & $84.20 \pm .17$ & $102.41 \pm .04$ & $4.47 \pm .05$          & $111.58 \pm .38$ \\
Sylvester   & $83.32 \pm .06$ & $99.00 \pm .04$  & $4.45 \pm .04$          & $104.62 \pm .29$ \\
\methodname & $\mathbf{82.82 \pm .01}$ & $\mathbf{98.33 \pm .09}$ & $\mathbf{4.39 \pm .01}$ & $\mathbf{104.03 \pm .43}$ 
\end{tabular}
\caption{Negative ELBO on test data for VAE models; \textbf{lower is better}. In nats for all datasets except Frey Faces which is presented in bits per dimension. Mean/stdev are estimated over 3 runs.}
\label{tab:inference}
\end{table}

\subsection{Variational Autoencoder}
We compare \methodname to other normalizing flows for use in variational inference.
We train a VAE~\citep{kingma2013auto} on four datasets using a \methodname flow and compare to VAEs with no flow, Planar Flows \citep{rezende2015variational}, Inverse Autoregressive Flow (IAF)~\citep{kingma2016improved}, and Sylvester normalizing flows \citep{berg2018sylvester}.
To provide a fair comparison, our encoder/decoder architectures and learning setup exactly mirror those of \citet{berg2018sylvester}.

In VAEs it is common for the encoder network to also output the parameters of the flow as a function of the input $\x$. With \methodname{}, we found this led to differential equations which were too difficult to integrate numerically.
Instead, the encoder network outputs a low-rank update to a global weight matrix and an input-dependent bias vector. 
Neural network layers inside of \methodname{} take the form
\begin{equation}
    \text{layer}(h; \x, W, b) = \sigma \left(\left(\underbrace{W \vphantom{\hat{U}(\x)}}_{\footnotesize D_{out} \times D_{in}} + \underbrace{\hat{U}(\x)}_{\footnotesize D_{out} \times k}{\underbrace{\hat{V}(\x)}_{D_{in} \times k}}^T\right) h + \underbrace{b \vphantom{\hat{U}(\x)}}_{D_{out} \times 1} + \underbrace{\hat{b}(\x) \vphantom{\hat{U}(\x)}}_{D_{out} \times 1} \right)
\end{equation}
where $h$ is the input to the layer, $\sigma$ is an element-wise activation function, $D_{in}$ and $D_{out}$ are the input and output dimensionality of this layer, and $\hat{U}(\x)$, $\hat{V}(\x)$, $\hat{b}(\x)$ are data-dependent parameters returned from the encoder networks.
A full description of the model architectures used and our experimental setup can be found in Appendix \ref{sec:vaeexperiment}.

On every dataset tested, \methodname outperforms all other competing normalizing flows. A summary of our variational inference results can be found in Table \ref{tab:inference}.

\section{Analysis and Discussion}
\begin{wrapfigure}[11]{r}{0.4\linewidth}
\vspace{-25mm}
\begin{center}
    \includegraphics[width=\linewidth]{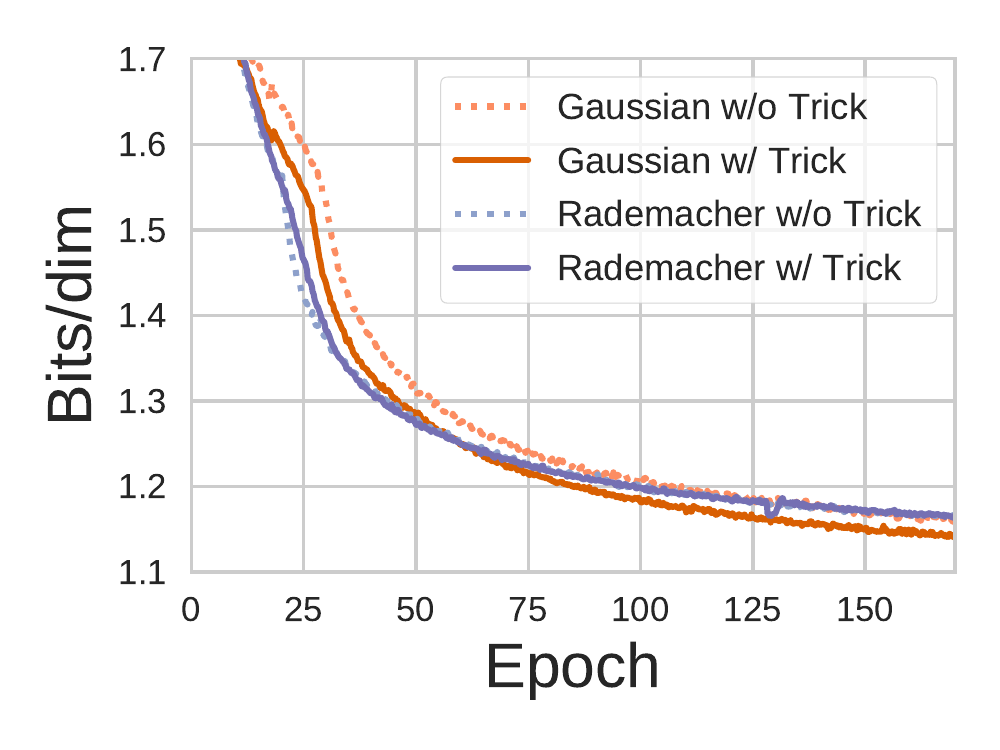}
\end{center}
\vspace{-4mm}
\caption{The variance of our model's log-density estimator can be reduced using neural network architectures with a bottleneck layer, speeding up training.}
\label{fig:bottleneck}
\end{wrapfigure}
We perform a series of ablation experiments to gain a better understanding of the proposed model. 
\subsection{Faster Training with Bottleneck Trick}

We plot the training losses on MNIST using an encoder-decoder architecture (see Appendix \ref{sec:densityexperiment} for details). 
Loss during training is plotted in Figure \ref{fig:bottleneck}, where we use the trace estimator directly on the $D\times D$ Jacobian or we use the bottleneck trick to reduce the dimension to $H \times H$. Interestingly, we find that while the bottleneck trick~\eqref{eq:bottleneck_trick} can lead to faster convergence when the trace is estimated using a Gaussian-distributed $\veps$, we did not observe faster convergence when using a Rademacher-distributed $\veps$. 

\subsection{Number of Function Evaluations vs.\\ Data Dimension}
\begin{wrapfigure}[8]{r}{0.4\linewidth}
\vspace{-35mm}
\begin{center}
    \includegraphics[width=\linewidth]{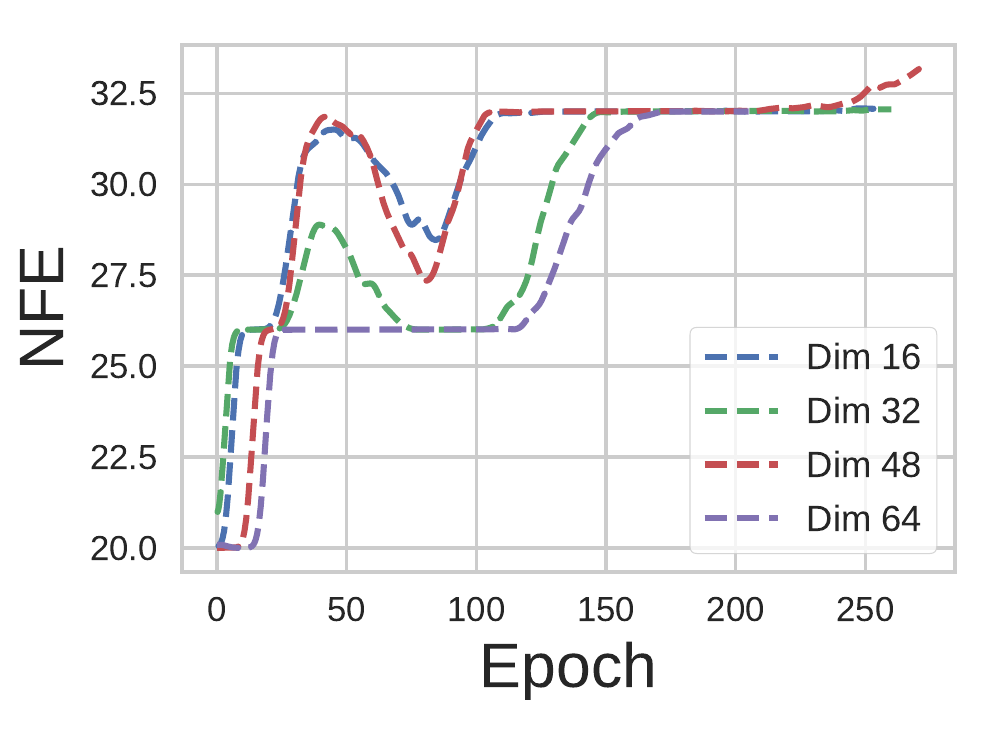}
\end{center}
\vspace{-4mm}
\caption{Number of function evaluates used by the adaptive ODE solver (NFE) is approximately independent of data-dimension.}
\label{fig:nfe}
\end{wrapfigure}
The full computational cost of integrating the instantaneous change of variables \eqref{eq:cont-cov} is $\bigO(DH\widehat{L})$ where $D$ is dimensionality of the data, $H$ is the size of the hidden state, and $\widehat{L}$ is the number of function evaluations (NFE) that the adaptive solver uses to integrate the ODE. In general, each evaluation of the model is $\bigO(DH)$ and in practice, $H$ is typically chosen to be close to $D$. Since the general form of the discrete change of variables equation \eqref{eq:discrete_cov} requires $\bigO(D^3)$-cost, one may wonder whether the number of evaluations $\widehat{L}$ depends on $D$. 

We train VAEs using \methodname flows with increasing latent dimension $D$. The NFE throughout training is shown in Figure \ref{fig:nfe}.
In all models, we find that the NFE increases throughout training, but converges to the same value, independent of $D$.
This phenomenon can be verified with a simple thought experiment.
Take an $\R^D$ isotropic Gaussian distribution as the data distribution and set the base distribution of our model to be an isotropic Gaussian.
Then the optimal differential equation is zero for any $D$, and the number evaluations is zero.
We can conclude that the number of evaluations is not dependent on the dimensionality of the data but the complexity of its distribution, or more specifically, how difficult it is to transform its density into the base distribution.

\subsection{Single-scale vs. Multi-scale \methodname}

\begin{wrapfigure}[11]{r}{0.4\linewidth}
\vspace{-16mm}
\begin{center}
    \includegraphics[width=\linewidth]{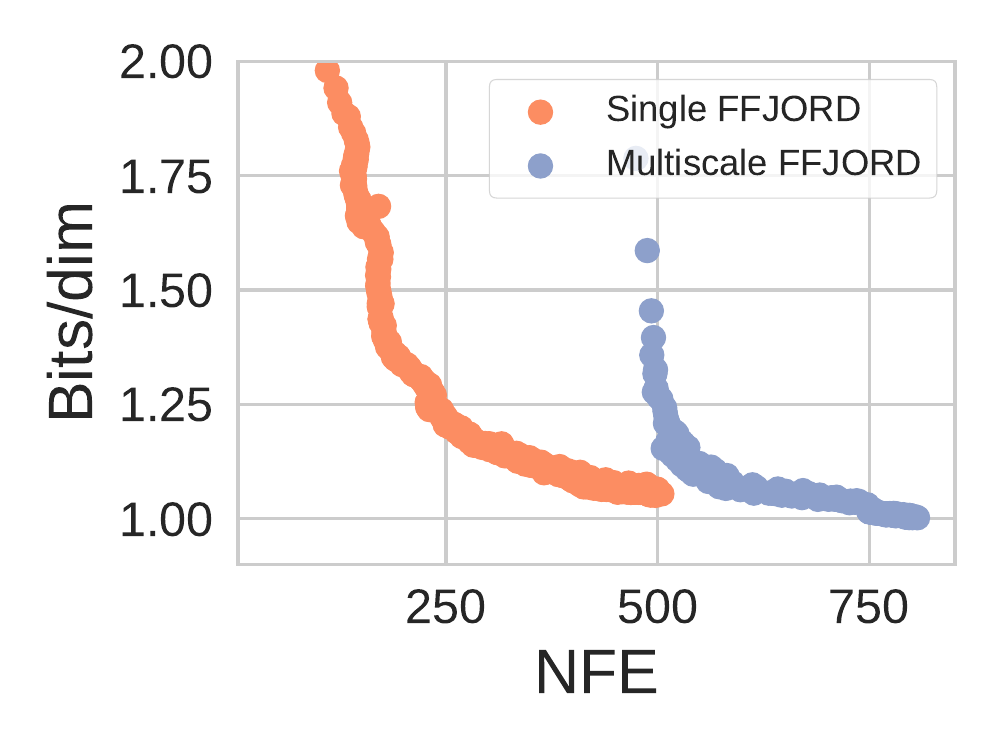}
\end{center}
\vspace{-4mm}
\caption{For image data, multiscale architectures require the ODE solver to use a greater number of function evaluations (NFE), but these models achieve better performance. }
\label{fig:nfescale}
\end{wrapfigure}

Crucial to the scalability of Real NVP and Glow is the multiscale architecture originally proposed in \cite{dinh2016density}. We compare an single-scale encoder-decoder style \methodname with a multiscale \methodname on the MNIST dataset where both models have a comparable number of parameters and plot the total NFE--in both forward and backward passes--against the loss achieved in Figure \ref{fig:nfescale}. We find that while the single-scale model uses approximately one half as many function evaluations as the multiscale model, it is not able to achieve the same performance as the multiscale model. 


\section{Scope and Limitations}



\paragraph{Number of function evaluations can be prohibitive.} The number of function evaluations required to integrate the dynamics is not fixed ahead of time and is a function of the data, model architecture, and model parameters.
We find that this tends to grow as the models trains and can become prohibitively large, even when memory stays constant due to the adjoint method.
Various forms of regularization such as weight decay and spectral normalization~\citep{miyato2018spectral} can be used to reduce the this quantity but their use tends to hurt performance slightly. 

\paragraph{Limitations of general-purpose ODE solvers.} In theory, we can model any differential equation (given mild assumptions based on existence and uniqueness of the solution), but in practice our reliance on general-purpose ODE solvers restricts us to non-stiff differential equations that can be efficiently solved.
ODE solvers for stiff dynamics exist, but they evaluate $f$ many more times to achieve the same error. 
We find that using a small amount of weight decay sufficiently constrains the ODE to be non-stiff.

\section{Conclusion}
We have presented \methodname, a reversible generative model for high dimensional data which can compute exact log-likelihoods and can be sampled from efficiently.
Our model uses continuous-time dynamics to produce a generative model which is parameterized by an unrestricted neural network.
All required quantities for training and sampling can be computed using automatic-differentiation, Hutchinson's trace estimator, and black-box ODE solvers.
Our model stands in contrast to other methods with similar properties which rely on restricted, hand-engineered neural network architectures.
We have demonstrated that this additional flexibility allows our approach to achieve improved performance on density estimation and variational inference. We also demonstrate \methodname's ability to model distributions which comparable methods such as Glow and Real NVP cannot model. 

We believe there is much room for further work exploring and improving this method. We are interested specifically in ways to reduce the number of function evaluations used by the ODE-solver without hurting predictive performance. Advancements like these will be crucial in scaling this method to even higher-dimensional datasets.  

\section{Acknowledgements}
We thank Roger Grosse and Yulia Rubanova for helpful discussions. 

\bibliography{iclr2019_conference}
\bibliographystyle{iclr2019_conference}

\clearpage
\begin{appendices}

\section{Qualitative Samples}\label{sec:imgen}
\begin{figure}
    \centering
    \begin{subfigure}[b]{0.05\linewidth}
    \rotatebox[origin=c]{90}{Samples}\vspace{16mm}
    \caption*{}
    \end{subfigure}
    \begin{subfigure}[b]{.45\linewidth}
    \centering
    \includegraphics[width=\linewidth]{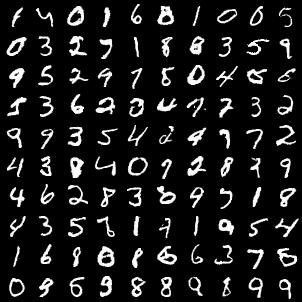}
    \end{subfigure}%
    \hfill%
    \begin{subfigure}[b]{.45\linewidth}
    \centering
    \includegraphics[width=\linewidth]{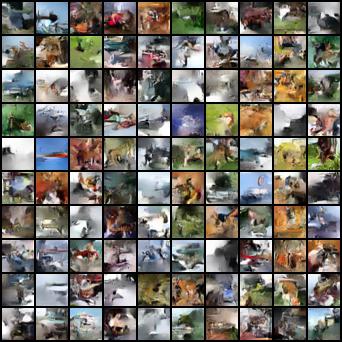}
    \end{subfigure}\\
    
    \begin{subfigure}[b]{0.05\linewidth}
    \rotatebox[origin=c]{90}{Data}\vspace{20mm}
    \caption*{}
    \end{subfigure}
    \begin{subfigure}[b]{.45\linewidth}
    \centering
    \includegraphics[width=\linewidth]{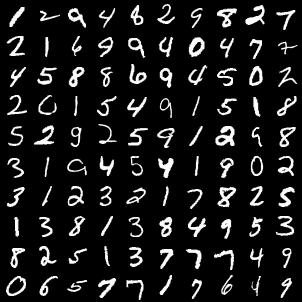}
    \end{subfigure}%
    \hfill%
    \begin{subfigure}[b]{.45\linewidth}
    \centering
    \includegraphics[width=\linewidth]{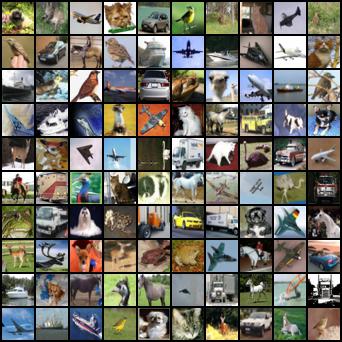}
    \end{subfigure}
    \caption{Samples and data from our image models. MNIST on left, CIFAR10 on right.}
    \label{fig:imgen}
\end{figure}

Samples from our \methodname models trained on MNIST and CIFAR10 can be found in Figure \ref{fig:imgen}.

\section{Experimental Details and Additional Results}
\subsection{Density Estimation}
\label{sec:densityexperiment}
On the tabular datasets we performed a grid-search over network architectures. We searched over models with 1, 2, 5, or 10 flows with 1, 2, 3, or 4 hidden layers per flow. Since each dataset has a different number of dimensions, we searched over hidden dimensions equal to 5, 10, or 20 times the data dimension (hidden dimension multiplier in Table \ref{tab:tabulararch}). We tried both the tanh and softplus nonlinearities. The best performing models can be found in the Table \ref{tab:tabulararch}.

On the image datasets we experimented with two different model architectures; a single flow with an encoder-decoder style architecture and a multiscale architecture composed of multiple flows. 

While they were able to fit MNIST and obtain competitive performance, the encoder-decoder architectures were unable to fit more complicated image datasets such as CIFAR10 and Street View House Numbers. The architecture for MNIST which obtained the results in Table \ref{tab:density} was composed of four convolutional layers with $64 \rightarrow 64 \rightarrow 128 \rightarrow 128$ filters and down-sampling with strided convolutions by two every other layer. There are then four transpose-convolutional layers who's filters mirror the first four layers and up-sample by two every other layer. The softplus activation function is used in every layer. 

The multiscale architectures were inspired by those presented in \cite{dinh2016density}. We compose multiple flows together interspersed with ``squeeze'' operations which down-sample the spatial resolution of the images and increase the number of channels. These operations are stacked into a ``scale block'' which contains $N$ flows, a squeeze, then $N$ flows. For MNIST we use 3 scale blocks and for CIFAR10 we use 4 scale blocks and let $N = 2$ for both datasets. Each flow is defined by 3 convolutional layers with 64 filters and a kernel size of 3. The softplus nonlinearity is used in all layers.

Both models were trained with the Adam optimizer \citep{kingma2014adam}. We trained for 500 epochs with a learning rate of .001 which was decayed to .0001 after 250 epochs. Training took place on six GPUs and completed after approximately five days. 

\subsection{Variational Autoencoder}
\label{sec:vaeexperiment}
Our experimental procedure exactly mirrors that of \cite{berg2018sylvester}.
We use the same 7-layer encoder and decoder, learning rate (.001), optimizer (Adam~\cite{kingma2014adam}), batch size (100), and early stopping procedure (stop after 100 epochs of no validaiton improvment).
The only difference was in the nomralizing flow used in the approximate posterior.

We performed a grid-search over neural network architectures for the dynamics of \methodname.
We searched over networks with 1 and 2 hidden layers and hidden dimension 512, 1024, and 2048.
We used flows with 1, 2, or 5 steps and wight matrix updates of rank 1, 20, and 64.
We use the softplus activation function for all datasets except for Caltech Silhouettes where we used tanh. The best performing models can be found in the Table \ref{tab:vaearch}.
Models were trained on a single GPU and training took between four hours and three days depending on the dataset.   

\begin{table}[H]\centering
\renewcommand{\arraystretch}{1.3}
\begin{tabular}{l|ccccc}
Dataset & nonlinearity & \# layers & hidden dim multiplier & \# flow steps & batchsize \\
\midrule
POWER     & tanh     & 3 & 10 & 5  & 10000 \\
GAS       & tanh     & 3 & 20 & 5  & 1000 \\
HEPMASS   & softplus & 2 & 10 & 10 & 10000 \\
MINIBOONE & softplus & 2 & 20 & 1  & 1000 \\
BSDS300   & softplus & 3 & 20 & 2  & 10000
\end{tabular}
\caption{Best performing model architectures for density estimation on tabular data with \methodname.}
\label{tab:tabulararch}
\end{table}

\begin{table}[H]\centering
\renewcommand{\arraystretch}{1.3}
\begin{tabular}{l|ccccc}
Dataset & nonlinearity & \# layers & hidden dimension & \# flow steps & rank \\
\midrule
MNIST      & softplus & 2 & 1024 & 2 & 64\\
Omniglot   & softplus & 2 & 512  & 5 & 20\\
Frey Faces & softplus & 2 & 512  & 2 & 20\\
Caltech    & tanh     & 1 & 2048 & 1 & 20
\end{tabular}
\caption{Best performing model architectures for VAEs with \methodname.}
\label{tab:vaearch}
\end{table}

\subsection{Standard Deviations for Tabular Density Estimation}\label{sec:tabular_stdev}

\begin{table}[H]\centering
\setlength{\tabcolsep}{4pt}
\renewcommand{\arraystretch}{1.3}
\begin{tabular}{@{}l|ccccc@{}}
& POWER & GAS & HEPMASS & MINIBOONE & BSDS300  \\
\midrule
Real NVP & -0.17 $\pm$ 0.01 & -8.33 $\pm$ 0.14 & 18.71 $\pm$ 0.02 & 13.55 $\pm$ 0.49 & -153.28 $\pm$ 1.78 \\
Glow & -0.17 $\pm$ 0.01 & -8.15 $\pm$ 0.40 & 18.92 $\pm$ 0.08 & 11.35 $\pm$ 0.07 & -155.07 $\pm$ 0.03 \\
\methodname & -0.46 $\pm$ 0.01 & -8.59 $\pm$ 0.12 & 14.92 $\pm$ 0.08 & 10.43 $\pm$ 0.04 & -157.40 $\pm$ 0.19 \\
\midrule
MADE & 3.08 $\pm$ 0.03 & -3.56 $\pm$ 0.04 & 20.98 $\pm$ 0.02 & 15.59 $\pm$ 0.50 & -148.85 $\pm$ 0.28 \\
MAF  & -0.24 $\pm$ 0.01 & -10.08 $\pm$ 0.02 & 17.70 $\pm$ 0.02 & 11.75 $\pm$ 0.44 & -155.69 $\pm$ 0.28 \\
TAN & -0.48 $\pm$ 0.01 & -11.19 $\pm$ 0.02 & 15.12 $\pm$ 0.02 & 11.01 $\pm$ 0.48 & -157.03 $\pm$ 0.07 \\
MAF-DDSF & -0.62 $\pm$ 0.01 & -11.96 $\pm$ 0.33 & 15.09 $\pm$ 0.40 & 8.86 $\pm$ 0.15 & -157.73 $\pm$ 0.04 
\end{tabular}
\caption{Negative log-likehood on test data for density estimation models. Means/stdev over 3 runs.}
\label{tab:density_stdev}
\end{table}

\section{Numerical Error from the ODE Solver}\label{sec:numerical_error}
ODE solvers are numerical integration methods so there is error inherent in their outputs.
Adaptive solvers (like those used in all of our experiments) attempt to predict the errors that they accrue and modify their step-size to reduce their error below a user set tolerance.
It is important to be aware of this error when we use these solvers for density estimation as the solver outputs the density that we report and compare with other methods.
When tolerance is too low, we run into machine precision errors.
Similarly when tolerance is too high, errors are large, our training objective becomes biased and we can run into divergent training dynamics. 

\begin{figure}[H]
    \centering
    \includegraphics[width=0.48\linewidth]{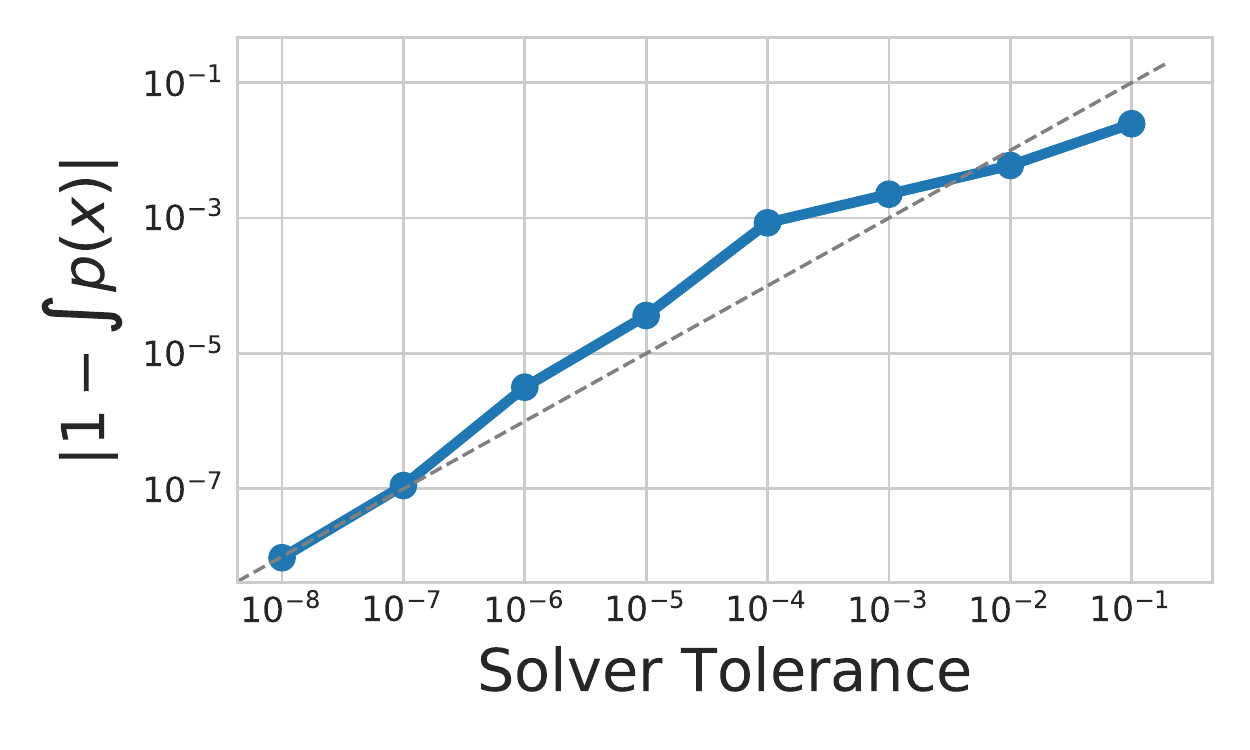}
    \includegraphics[width=0.48\linewidth]{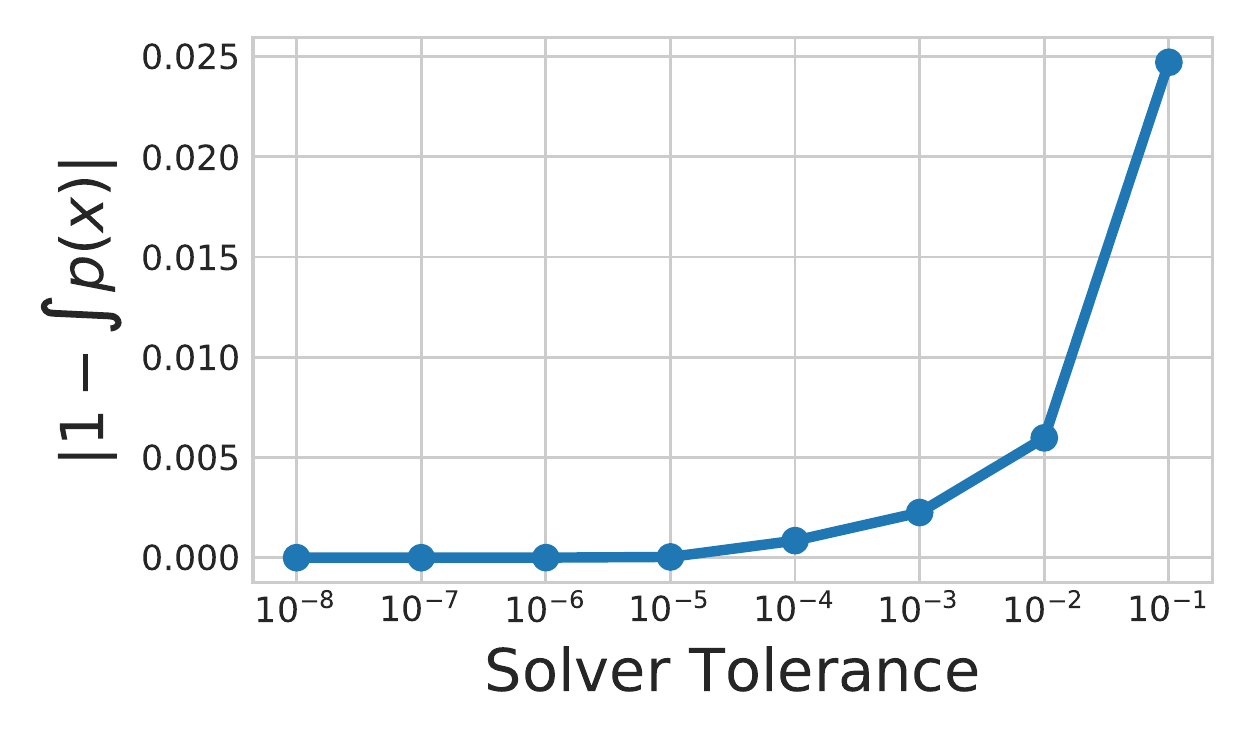}
    \caption{Numerical integration shows that the density under the model does integrate to one given sufficiently low tolerance. Both log and non-log plots are shown.}
    \label{fig:err_vs_tol}
\end{figure}

Since a valid probability density function integrates to one, we take a model trained on Figure \ref{fig:toy1d} and numerically find the area under the curve using Riemann sum and a very fine grid. We do this for a range of tolerance values and show the resulting error in Figure \ref{fig:err_vs_tol}. We set both \verb|atol| and \verb|rtol| to the same tolerance.

The numerical error follows the same order as the tolerance, as expected. During training, we find that the error becomes non-negligible when using tolerance values higher than $10^{-5}$. For most of our experiments, we set tolerance to $10^{-5}$ as that gives reasonable performance while requiring few number of evaluations. For the tabular experiments, we use \verb|atol|=$10^{-8}$ and \verb|rtol|=$10^{-6}$.

\end{appendices}

\end{document}